# AN INNOVATIVE SKIN DETECTION APPROACH USING COLOR BASED IMAGE RETRIEVAL TECHNIQUE

Shervan Fekri-Ershad[1], Mohammad Saberi[2] and Farshad Tajeripour[3]

[1,2,3]Dept. Computer Science & Engineering and IT, Shiraz University
Shiraz/ Iran
`shfekri@shirazu.ac.ir`
`m_saberi@shirazu.ac.ir`
`tajeri@shirazu.ac.ir`

## *ABSTRACT*

*From The late 90[th], "Skin Detection" becomes one of the major problems in image processing. If "Skin Detection" will be done in high accuracy, it can be used in many cases as face recognition, Human Tracking and etc. Until now so many methods were presented for solving this problem. In most of these methods, color space was used to extract feature vector for classifying pixels, but the most of them have not good accuracy in detecting types of skin. The proposed approach in this paper is based on "Color based image retrieval" (CBIR) technique. In this method, first by means of CBIR method and image tiling and considering the relation between pixel and its neighbors, a feature vector would be defined and then with using a training step, detecting the skin in the test stage. The result shows that the presenting approach, in addition to its high accuracy in detecting type of skin, has no sensitivity to illumination intensity and moving face orientation.*

## I. INTRODUCTION

Skin detection is one of the basic subjects in image processing. In many cases such as human detection and tracking, visual identification and face detection, a skin detection stage is needed. The concept of "skin detection" in an image is the classification of the existence pixels in that image into two skin and Non-skin classes. In this direction, several methods have been presented until now. In most of the proposed methods, researchers have tried to define and extract a feature vector for each pixel of image and in the end, classify the feature vectors.

In most methods, color space has been used to extract the features. For instance, in [1] RGB space, in [2] HSV space and in [3] $YC_bC_r$ space have been used. Also other color spaces like HSI, UCS have been used in some of the approaches. In some methods, also the texture analysis [4] and Mixture of Gaussians (MOG) [5] have been used to extract the more meaningful and necessary features. About skin detection problem, there are some major motivations that all methods try to solve or remove them. Such as:

a) Very low false rejection rate at low false detection rate
b) Detection of different skin color types and kinds
c) Handling of ambiguity between skin and non-skin colors
d) Robustness to variation in lighting conditions
e) Insensitivity to noise and face orientation

In this article, an approach based on color based image retrieval (CBIR) technique has been presented to resolve these motivations. In this approach, firstly a set of features is defined by CBIR technique and histogram analysis, and then by tiling the image and using a train level, a good threshold for classifying the pixels would be achieved. The given approach in this article has the ability of detecting all kinds of skin because of using the train level. Also because of considering the relation of every pixel with its neighbors, it's not sensitive to noise, illumination and changing the orientation of face or body. The results confirm this claim. It is mentionable that the given method in this article has a general aspect and in other cases in which the aim is two class classification, it also can be used like defect detection.

## II. IMAGE RETRIEVAL TECHNIQUE

Image retrieval is one of the efficient and new problems in image processing area. The main aim in image retrieval is retrieving the most similar images to the query image from a huge database. In the direction of achieving this aim, researchers try to define the most effective and meaningful features to compare images. Yet, various methods have been proposed for image retrieval. In [6] a method based on energy compaction has been given. In a recent work Ershad [7] described a method based on primitive pattern units. In [8] some features based on texture analysis has been given. Vasconcelos [9] has proposed a method according to the features of shape. In a large set of methods presented so far, color histogram analysis of image has been used to define features.

One of the approaches which had some desirable results is color based image retrieval (CBIR) [10]. In this method for comparing the similarity rate of two image, firstly the histogram of two images would be individually extracted in each color channel Red, green and blue and then compared by a criterion like Euclidean one [11]. Equation (1) represents this affair.

$$d = \sqrt{\sum_{i=1}^{n} |H1_i - H2_i|^2} \qquad (1)$$

Where, H1 and H2 are the histogram of image (1) and image (2), and "i" means the number of histogram bins. Also, "d" is the Euclidean distance between first and second images, which can compute in each color spaces(R, G or B) individually.

## II.I. PROPOSED FEATURE EXTRACTION

In this article, this technique has been inspired to define feature vector. In this way that first the histogram of image would be individually extracted and normalized in each color channels such as red, green and blue. Then every bin of histogram would be considered as a dimension of feature vector and the bin's height is seen as the value of that dimension. So we have a feature vector like F including 256*3 dimensions. The experimental results conclusions that vector F, as defined, could not provide a good discrimination and severability. The reason is the very small values of vector F in every dimension. So vector "F" would be quantized and changed into F′ form. In F′ vector, the value of each dimension is the sum of N dimension values from vector F. so for example if N=16 then F′ vector has 16*3 dimensions. In equation (2) and (3), F and F′ vectors are represented. Considering the above descriptions, F′ vector can be calculated for every image or slice of color image. F′ would be a good identification for that image.

$$F = <H_{i1}, H_{i2}, \ldots, H_{i256}> \quad (2)$$

$$F' = <\sum_{i=1}^{N} H_i, \sum_{i=N+1}^{2N} H_i, \ldots, \sum_{i=256-N}^{256} H_i> \quad (3)$$

Where, H is the histogram of image in each channel. So, "i" means the Red, Green and Blue channels which all of them should compute. Also, N is the range of quantization.

## III. PROPOSED SKIN DETECTION APPROACH

According to the explanations of section2, we can define a feature vector like F′ for every image, which is a good identification of that image. So we can also use F′ to detect skin. In order that F′ vector be a good identification of skin image, a train stage is needed.

In train stage, first some images including just skin would be prepared. These images are called *Pure Skin* images. Then by the method described in section2, F′ vector is calculated for each train images. Finally, by equation (4), the average of the calculated feature vectors would be computed. Now, the average vector is a good identification for skin.

$$F_{Average} = \sum_{j=1}^{M} F'_j \quad (4)$$

Where, M is the number of train images and F′ is the feature vector that computed for $M_{th}$ image. The most methods, presented for skin detection until now, don't consider the relation between each pixel with its neighborhoods and this decrease their accuracy rate. In our approach, in order to consider the relation between each pixel with its neighborhoods, in test stage, we tilt the image into windows with equal dimensions and without overlapping. This act called image tiling [10], see figure (1).

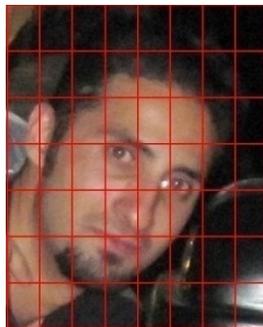

Figure(1). Description of image tiling

The International Journal of Multimedia & Its Applications (IJMA) Vol.4, No.3, June 2012

Now after image tiling, in test stage, feature vector is computed for each window, according to section 2. Then by computing the distance between the calculated feature vector for each window and average vector, and after that, comparing this distance by skin threshold , it could be determined that which window has pixels of skin and which one doesn't. Equation (5) shows this.

$$D = distance\ (F_{image}, F_{Average})$$

$$Skin\ Detection = \begin{cases} Skin & if\ D \leq T \\ Non - Skin & else \end{cases} \quad (5)$$

Where, "D" is the distance amount and "distance" means the used criteria for similarity. Also, the "T" is threshold for skin. The threshold tuning method is described in next section.

For comparing the distance between two vectors, there are a lot of criteria such as City Block, Minkowski, Sorensen, Chebyshev and etc [12]. The results show that the best distance criterion for this approach is "Gower"[12]. This criterion has been shown in equation (6). Also, the quality of determining the optimized threshold has been explained in section 4.

$$D = \frac{1}{N} \sum_{i=1}^{N} \frac{|F_{1i} - F_{2i}|}{R_i} \quad (6)$$

Where, $F_1$ and $F_2$ are the computed feature vectors and "i" means the $i_{th}$ dimensions of feature vector. Also, N is the size of feature vector and D means the distance amount. Also R means the range of $i_{th}$ dimension. The flowchart of proposed approach is shown in figure (2). To prove and compare the results of Gower distance by other criteria, in the result part the quality of Bhattacharyya (Eq 7), City Block( Eq 8) and Soergel (Eq 9) distance are computed. These are shown in the table 2.

$$D(\text{Bhattacharyya}) = -ln \sum_{i=1}^{N} \sqrt{F_{1i}F_{2i}} \quad (7)$$

$$D(City\ Block) = \sum_{i=1}^{N} |F_{1i} - F_{2i}| \quad (8)$$

$$D(Soergel) = \frac{\sum_{i=1}^{N}|F_{1i} - F_{2i}|}{\sum_{i=1}^{N} Max(F_{1i}, F_{2i})} \quad (9)$$

60

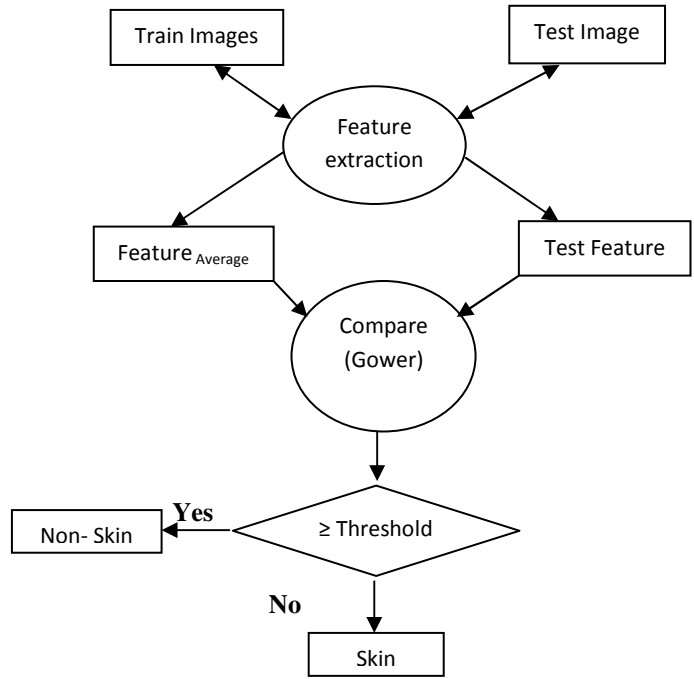

Figure(2). Flowchart of proposed approach

## IV. THRESHOLD TUNING

As it was mentioned in equation (5), a proper threshold is necessary to recognize being skin of every window. We can determine the optimized threshold by a train level. As we said in section3, pure skin images are used to compute average vector. Thus for threshold, firstly every pure skin image of train stage is tilted into windows with the same dimensions of the windows in test stage. Then for every window a feature vector is extracted. After that the distance of every vector is measured with average vector by equation (6). Finally, the greatest obtained distance is considered as skin threshold.

### IV.I. THRESHOLD TUNING FOR EVERY KIND OF SKINS

As mentioned in the introduction, one of the advantages of the given approach in this paper is its ability to detect different kinds of skin. In order that this approach distinguish different kinds of skin such as yellow, white, red and etc, it's enough to provide a set of pure skin images from every kind. Then, in train level, an average feature vector should be extracted for every kind of skin. In test stage, at first an optimized threshold should be gained for each kind of skin, considering the given algorithm in section4. Finally, the distance vector of that window with all average vectors should be determined for every window of test. If at least one of the distances is less than threshold related to the kind of skin, that window is skin, otherwise it isn't skin. Equation (10) shows this.

$$Skin\ \&\ color\ Detection = \begin{cases} Skin\ by\ Color(i) & if\ D_i \leq T_i \\ Non-Skin & else \end{cases} \quad (10)$$

Where, T is computed threshold and "i" means kinds of skin like red, white, black or etc. Also, $D_i$ is the computed distance between windows feature vector and average vector of skin "i".

## V. RESULTS

In this section to research the quality of the proposed approach, 50 images of human in different conditions, from the orientation of face and body view, illumination and background were provided. In these images, there were 3 kinds of skin such as red, yellow and white. Also for train stage, 10 pure skin images were provided from every kind of skin. So the written program by Matlab software was trained by the help of them. After that the approach was tested with different sizes for image tiling. Finally the best conclusions were obtained for 16*16 dimensions. To study the quality of the given approach, equation (11) was used [13, 16]. Equation (11) has been used in most related to defect detection and skin detect. The accuracy rate of the approach was computed for every 50 images. Their average accuracy rate was 93.47 ± 0.6. The standard deviation was determined by "Pair T.Test" algorithm. Also some of previous approaches such as [14] and [15] are applied on results images and compared with proposed approach in table 1. Some of the results are shown in figure (4).

$$Detection\ Rate = 100 \times \frac{N_{cc} + N_{dd}}{N_{total}} \qquad (11)$$

Where, $N_{cc}$ shows the number of windows which are really non-skin and the proposed approach detected it as non-skin window. And the $N_{dd}$ means the windows which detected as skin in real image it is skin. To determining and counting Skin or Non-skin windows, the Skin detected image is divided to non-overlapping windows with same size by train windows. Next, each window that includes just one Skin detected pixel categorized as Skin window. There is a same way for Non-Skin windows. It is shown in figure (3). For example, the window which includes in third column and fifth row is categorized as skin window.

The Sensitivity (Eq 12) and specificity (Eq 13) were measured for all of the skin detected results. Where TN, TP, FN and FP are means true negative, true positive, false negative and false positive. These are shown in table 1.

To prove and compare the results of Gower distance by other criteria, in the result part the quality of Bhattacharyya (Eq 7), City Block( Eq 8) and Soergel (Eq 9) distance are computed. These are shown in the table 2.

$$Sensitivity = \frac{TP}{TP + FN} \qquad (12)$$

$$Specificity = \frac{TN}{TN + FP} \qquad (13)$$

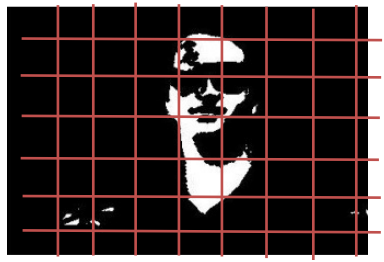

Figure (3). Determining result by skin detected image

Table1. Accuracy rate of proposed approach and some other methods

| Skin detection methods | Accuracy rate ± std | Sensitivity | Specificity |
|---|---|---|---|
| *Proposed approach* | **93.47 ± 0.6** | **92.28** | **95.34** |
| *Ref.[14]* | 88.62 ± 0.8 | 85.32 | 89.01 |
| *Ref.[15]* | 81.22 ± 0.9 | 79.54 | 82.81 |

Table2. Accuracy rate of proposed approach based on difference criteria

| Criteria | Gower | Bhattacharyya | City Block | Soergel |
|---|---|---|---|---|
| *Proposed approach* | **93.47 ± 0.6** | 91.12± 0.7 | 84.47± 2.2 | 89.06± 0.9 |

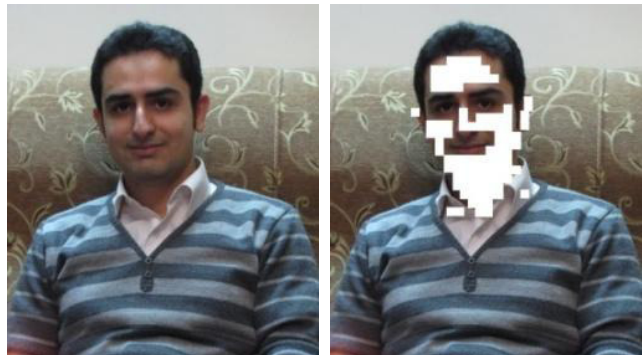

(a)          (b)

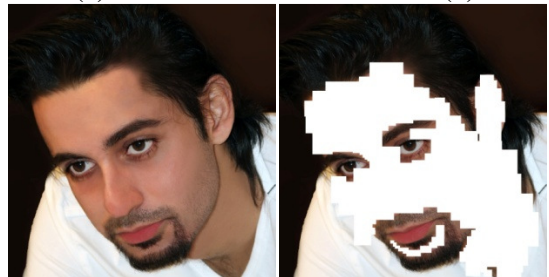

(c)          (d)

The International Journal of Multimedia & Its Applications (IJMA) Vol.4, No.3, June 2012

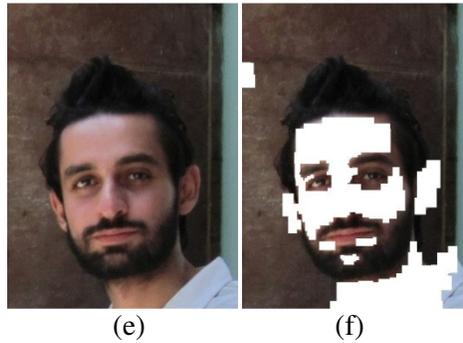

(e)                   (f)

Figure (4). Some of the experimental results
(a) Original image (b) Skin Detected of (a)
(c) Original Image (d) Skin Detected of (c)
(e) Original image (f) Skin Detected of (e)

## VI. CONCLUSION

The main aim of this article, was presenting an accurate approach for skin detection. In this regard, first some of the previous methods were analyzed, and then the approach of this paper proposed based on CBIR technique. The proposed approach contains two steps such as train and test. First in training step, pure skin images were trained and then in testing steps skin area were detected from non-skin areas. The results showed that the presented approach in detecting skin types has high accuracy. In comparison to some of the previous methods, the benefits of this method are:

a) Low computational complexity
b) Low time complexity
c) High accuracy for all kinds of skin
d) Capability of skin types classification.

In addition, this method cause of its training step can be used in many of visual multi class classification cases such as defect detection. Its worth nothing that accuracy of approach according to image background, can be decreased for some kinds of skin, of course this problem can be solved by training step improvement.

## VII. Acknowledgment

The authors like to say thanks to Mrs. Paria Abadgar (MS.c. of art, University of Isfahan,Iran) for Her English translation. Also, the authors like to say thanks with best wishes to Mrs. Sana Karimian Ardestani (BS.c. of  Agricultural Engineering-Plant medicine, Azad University of Jahrom, Iran) and Mrs. Bahar Abadgar (BS.c. of Art, University of Isfahan, Iran)  for their support.